\documentclass[runningheads]{llncs}
\usepackage{color}
\usepackage{graphicx}
\usepackage{amssymb}
\usepackage{amsmath}
\usepackage{booktabs}
\usepackage{xcolor}
\usepackage{tabularx} 
\usepackage{arydshln}
\usepackage{multirow}
\usepackage{makecell}
\usepackage{url}
\usepackage{hyperref}

\begin{document}
\title{Low-Rank Mixture-of-Experts for \\ Continual Medical Image Segmentation}
%
\author{Qian Chen\inst{1,2,3,4} \and
Lei Zhu\inst{1,2,3,4} \and
Hangzhou He\inst{1,2,3,4} \and
Xinliang Zhang\inst{1,2,3,4}\and
Shuang Zeng\inst{1,2,3,4}\and
Qiushi Ren\inst{1,2,3,4}\and
Yanye Lu\inst{1,2,3,4}}
%
%
\institute{Department of Biomedical Engineering, Peking University, Beijing, China \and Institute of Medical Technology, Peking University, Beijing, China 
\and Institute of Biomedical Engineering, Peking University Shenzhen Graduate School, Shenzhen, China
\and  National Biomedical Imaging Center, Peking University, Beijing, China}
%
\maketitle              
\begin{abstract}
The primary goal of continual learning (CL) task in medical image segmentation field is to solve the “catastrophic forgetting” problem, where the model totally forgets previously learned features when it is extended to new categories (class-level) or tasks (task-level). Due to the privacy protection, the historical data labels are inaccessible. Prevalent continual learning methods primarily focus on generating pseudo-labels for old datasets to force the model to memorize the learned features. However, the incorrect pseudo-labels may corrupt the learned feature and lead to a new problem that the better the model is trained on the old task, the poorer the model performs on the new tasks. To avoid this problem, we propose a network by introducing the data-specific Mixture of Experts (MoE) structure to handle the new tasks or categories, ensuring that the network parameters of previous tasks are unaffected or only minimally impacted. To further overcome the tremendous memory costs caused by introducing additional structures, we propose a Low-Rank strategy which significantly reduces memory cost. 
We validate our method on both class-level and task-level continual learning challenges. Extensive experiments on multiple datasets show our model outperforms all other methods.

\end{abstract}
\section{Introduction}

Medical image segmentation (MIS) has been extensively studied and is crucial for quantitative disease analysis \cite{iyer2016quantitative}, computer-aided diagnosis \cite{roth2015improving}, and cancer radiotherapy planning \cite{jin2021deeptarget}. The advancement of deep learning technologies has boosted the field. However, the current setting of deep learning scenarios has limitations compared to the practical deployment in clinical medical environments. Present mainstream models can only segment single-modal datasets of lesions or organs, yet actual clinical practitioners require dynamic extending to identify targets across multi-modal datasets. Additionally, in the scenario of multi-organ segmentation, there is also an expectation that segmentation models can dynamically segment new organs without accessing old datasets. This anticipated clinical scenario can be understood as a continual learning (CL) problem in MIS. Models are easily prone to forgetting old data while learning new knowledge. This problem, known as ``catastrophic forgetting'' \cite{lewandowsky1995catastrophic} in CL, is an urgent issue that needs to be addressed. It is worth noting that in CL, data arrives at the model in sequence, and when the model learns new knowledge, it has no access to the old dataset.

Current research also explores the issue of CL in MIS, where the most critical problem is how to prevent catastrophic forgetting. Common approaches are regularization-based methods, which primarily use pseudo-labeling for old classes. These methods \cite{zhang2023continual,liu2022learning} train on datasets from both old and new classes together to achieve test results on both. The issue with this approach is that the accuracy of pseudo-labels is not high, and since the model is mainly trained on new datasets, the performance on both old and new classes is not satisfactory. 
Architecture-based methods are dedicated to dynamically adding dataset-specific partial networks \cite{golkar2019continual,hung2019compacting} or expanding the network by fixing the parameters of old tasks and adding new parameters for new tasks \cite{rusu2016progressive,li2019learn}. Although the issue of catastrophic forgetting is addressed, it leads to tremendous memory costs for model parameters.

We summarize the ongoing issues in CL for MIS. (1) Can we design networks that avoid the problem of catastrophic forgetting? (2) There is a clear trade-off in the CL capability when handling new and old tasks, the better the model performs on old datasets, the worse it tends to perform on new datasets \cite{ji2023continual,li2017learning}. Can the model maintain the ability to learn continuously over multiple learning steps? (3) Whether it's applying pseudo-labels or adding extra parameters, both introduce significant costs. Can the model have the advantages of being both lightweight and cost-effective?

To address the aforementioned problems, we propose a novel network structure for CL of MIS, named the Low-rank Mixture of Experts (MoE) architecture. This is a Transformer-based network structure, where an MoE layer consists of \textit{E} feed-forward networks ${\rm FFN}_1 ... {\rm FFN}_E$. First, during each training session, we only update the dataset-specific low-rank expert and fix all other parts of the network. As new datasets arrive, we append another expert layer and proceed to train solely that particular expert. In doing so, the knowledge acquired from previous tasks is not lost since the parameters within their respective experts are already fixed. This approach effectively resolves the issue of catastrophic forgetting since the knowledge learned on old tasks and new tasks is preserved by different experts. Therefore, the model can maintain its ability for CL across different steps, achieving high accuracy in training on both old and new tasks. It also presents a significant drawback despite the clear advantages of this multi-expert model: incorporating multiple dataset-specific experts into the network substantially increases computational and parameter costs. To mitigate this, we adopt a low-rank strategy. The weights within the MoE FFN are decomposed into a dimension-reducing matrix B and a dimension-increasing matrix A. Throughout training, we maintain the original pretrained weights $W_0$ and update only the parameters within A and B. 

To validate the effectiveness of the Low-Rank MoE structure, we conduct two series of experiments in both task-level CL settings (cross-modal multi-dataset medical image segmentation task) and class-level CL settings (multi-organ segmentation task). Our proposed method are evaluated on five datasets: ACDC \cite{bernard2018deep}, ISIC \cite{codella2019skin}, COVID-19 Segmentation dataset \cite{fan2020inf}, BTCV \cite{landman2015miccai}, and LiTS \cite{bilic2023liver}. In the task-level CL setting, we find that the low-rank experts from earlier tasks can not only assist subsequent tasks but also resolve the catastrophic forgetting problem. In the class-level CL setting, we introduce a language-guided gating function that successfully achieves CL of multi-organ segmentation across new and old classes. Moreover, we compare our method with several popular regularization-based approaches \cite{li2017learning,michieli2019incremental,douillard2021plop,ji2023continual}. The thorough results demonstrate the effectiveness of our proposed Low-Rank MoE method in achieving CL for cross-modal multi-dataset segmentation task and multi-organ segmentation task.

\section{Methodology}
Let $\mathcal{X}$ be the input image space and $\mathcal{Y}$ be the label space. In the CL for MIS setting, the training procedure is arranged into multiple steps, and each learning step $\emph{t}$ will involve novel class $\mathcal{C}^t$, constructing a new label set $\mathcal{Y}^t = \mathcal{Y}^{t-1} \cup \mathcal{C}^t$. When training on the $t-$th dataset $D_t$, the previous datasets of $\{D_1,...,D_{t-1}\}$ are not seen. The model is required to predict the accumulated labels for all seen datasets $\{D_1,...,D_t\}$:
\begin{equation}
y_i=argmax_{c\in \mathcal{C}^t }P(y_i=c|\mathcal{X}),\mathcal{C}^t=\cup_{r\leq t}\mathcal{C}^r
\end{equation}
where $P$ is the probability function that the model learns and $y_i$ is the output mask.
\subsection{Low-Rank Mixture of Experts}

\begin{figure}
\centering
\includegraphics[width=0.8\textwidth]{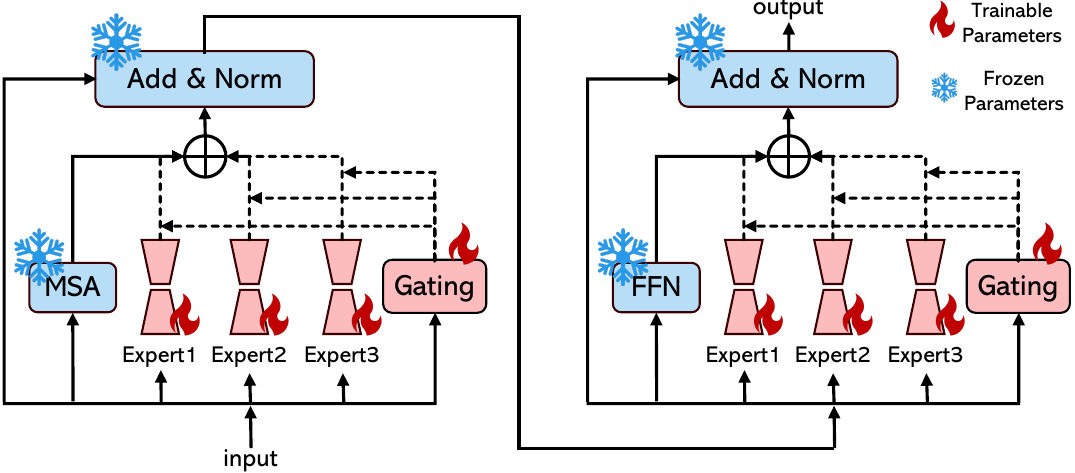}
\vspace{-1em}
\caption{An overview of Low-Rank MoE architecture. MSA means multi-head self-attention module.  }
\vspace{-1em} 
\label{fig1}
\end{figure}
\subsubsection{Low-Rank Mixture of Experts Layers}To facilitate CL in MIS, we employ a Mixture of Experts (MoE) design to achieve this. Figure~\ref{fig1} illustrates the overall framework of the proposed model architecture. We follow the Mixture-of-Experts Transformer models proposed by \cite{lepikhin2020gshard}. 
A MoE layer for Transformer consists of \textit{E} feed-forward networks (FFN) ${\rm FFN}_1 ... {\rm FFN}_E$.
$   {\rm FFN}_{e}(x_s)=Wo_e\cdot {\rm GeLU}(Wi_e\cdot x_s)$,
$y_s = \sum_{e=1}^E  G_{s,e}\cdot {\rm FFN_e}(x_s)$,
where $x_s$ is the input token at position $s$ to the MoE layer and each ${\rm FFN_e}$ is a two layer neural network using a GeLU activation function. $Wi_e$ and $Wo_e$ are the input and output projection weights of the $e$-th expert. Vector $ G_{s,E}$ is computed by a gating network. 
LoRA(Low-Rank Adapter) has been demonstrated to be an effective and efficient way to adapt pre-trained models to specific tasks\cite{hu2021lora}. Formally, for a pre-trained weight matrix $W_0\in \mathbb{R}^{d\times k}$, LoRA updates the $W$ with a low-rank decomposition: $W_0+\Delta W=W_0+BA$, $B \in \mathbb{R}^{d\times r}$, $A \in \mathbb{R}^{r\times k}$, and the rank $r\ll min(d,k)$. During training, $W_0$ is frozen and does not receive gradient updates, while $A$ and $B$ contain trainable parameters. We use a random Gaussian initialization for $A$ and zero for $B$, so $\Delta W =BA$ is zero at the beginning of training and doesn't affect the generalization ability of the pre-trained weights $W_0$. We use Low-Rank Adapters as the experts for different tasks and adapt them for the FFN layers and the attention layers. This strategy facilitates adaptive model updates for new tasks without losing valuable information learned in previous tasks. Specifically, the forward process of the LoRA FFN MoE layer can be formulated as:
\begin{equation}
   \begin{aligned}
    {\rm FFN}_{e}(x_s)&=(Wo+\Delta W^o_e)\cdot {\rm GeLU}((W^i+\Delta W^i_e)\cdot x_s)\label{equ.6}    
   \end{aligned}
\end{equation}
where $\Delta W^i_e$ and $\Delta W^o_e$ are the input and output low-rank projection weights of the $e$-th expert. 

\subsubsection{Low-Rank MoE Attention}
For the modified low-rank attention module, we replace four regular linear layers with low-rank linear layers. Formally, the matrix operation of the LoRA multi-head attention can be expressed as:
\begin{equation}
    \begin{aligned}
    MultiHead(Q,K,V)=\rm Concat(head_1,...,head_h)(W^O+B^O A^O)
    \end{aligned}
\end{equation}
where $W^O \in \mathbb{R}^{d_{model} \times d_{model}}$ and
\begin{equation}
   \rm head_i=Attention[Q(W_i^Q + B_i^Q A_i^Q), K(W_i^K + B_i^K A_i^K), V(W_i^V + B_i^V A_i^V)]
\end{equation}
where the projections are parameter matrices $W_i^Q \in \mathbb{R}^{d_{model} \times d_k}$, $B_i^Q \in \mathbb(R)^{d_{model}\times r}$, $A_i^Q \in \mathbb(R)^{r \times d_k}$; $W_i^K \in \mathbb{R}^{d_{model} \times d_k}$, $B_i^K \in \mathbb(R)^{d_{model}\times r}$,$A_i^K \in \mathbb(R)^{r \times d_k}$;
$W_i^V \in \mathbb{R}^{d_{model} \times d_k}$, $B_i^V \in \mathbb(R)^{d_{model}\times r}$, $A_i^V \in \mathbb(R)^{r \times d_k}$.
In this work, we employ $h=8$ parallel attention heads\cite{vaswani2017attention}, $d_k = d_v =d_{model}/h=64$, r=8\cite{hu2021lora}.
\subsection{Continual Learning Gating Strategy}
\subsubsection{Task-level Gating:} 

\begin{figure}
\centering
\includegraphics[width=0.9\textwidth]{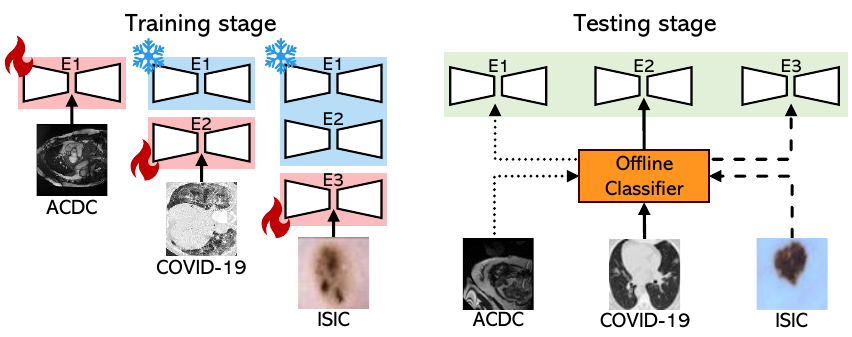}
\vspace{-1.5em}
\caption{Illustration of the proposed task-level gating pipeline.} 
\vspace{-1em}
\label{fig2}
\end{figure}

For task-level model, We utilize the popular SETR \cite{zheng2021rethinking} as the backbone. As illustrated in~\ref{fig2}, given a new task $T_k$ associated with its data $D_k$, we directly train an expert $E_k$ on $D_k$. For the segmentation task on each medical dataset, we learn a specialized expert for it. During step 1, dataset $D_1$ is processed by the first FFN, $FFN_{E_1}$. The matrix operation of the linear layer in the $FFN_{E_1}$ can be expressed as $h=W_{0}x+B_{E_1} A_{E_1} x$. In step 2, dataset $D_2$ is processed by $FFN_{E_2}$, which is superimposed on $FFN_{E_1}$. The matrix operation of the linear layer in the $FFN_{E_2}$ can be expressed as $h=W_{0}x+B_{E_1} A_{E_1} x+B_{E_2} A_{E_2} x$, the parameters in the $FFN_{E_1}$ are fixed, i.e. $B_{E_1} A_{E_1}$ are frozen. We can represent the matrix operation of the linear layer in the task $t$ in the $FFN_{E_t}$ as:
\begin{equation}
\begin{split}
    h &= W_{0}x + \sum_{t=1}^T B_{E_t} A_{E_t} x =\underbrace {(W_{0} + \sum_{t=1}^{T-1} B_{E_t} A_{E_t})}_{frozen} x + B_{E_T} A_{E_T} x\label{equ.10}
\end{split}
\end{equation}
Thus, according to the equation \ref{equ.6} and \ref{equ.10}, the forward process of the LoRA MoE layer in the $T_{th}$ task FFN layer can be represented as:
\begin{scriptsize}
\begin{equation}
    FFN_{E_T} = (\underbrace{W_0 +\sum_{t=1}^{T-1}B^o_{E_t} A^o_{E_t }}_{frozen}+B^o_{E_T} A^o_{E_T })\cdot GeLU((\underbrace{W^i +\sum_{t=1}^{T-1}B^i_{E_t} A^i_{E_t }}_{frozen}+B^i_{E_T} A^i_{E_T })\cdot x)
\end{equation}
\end{scriptsize}where $i$ and $o$ represent the input and output linear layer since there are two linear layers in one FFN layer. We freeze the weights corresponding to the old tasks and only update the weights in the new task.
When testing $T_k$,  we first input the test data $D_k$ into a scalable matching-based offline classifier (with high classification accuracy 99.7\%) to determine the task number, and then use the corresponding expert model $E_K$ to test $D_k$ according to the task number.
\subsubsection{Class-level Gating:}
\begin{figure}
\centering
\vspace{-1.5em}
\includegraphics[width=1.0\textwidth]{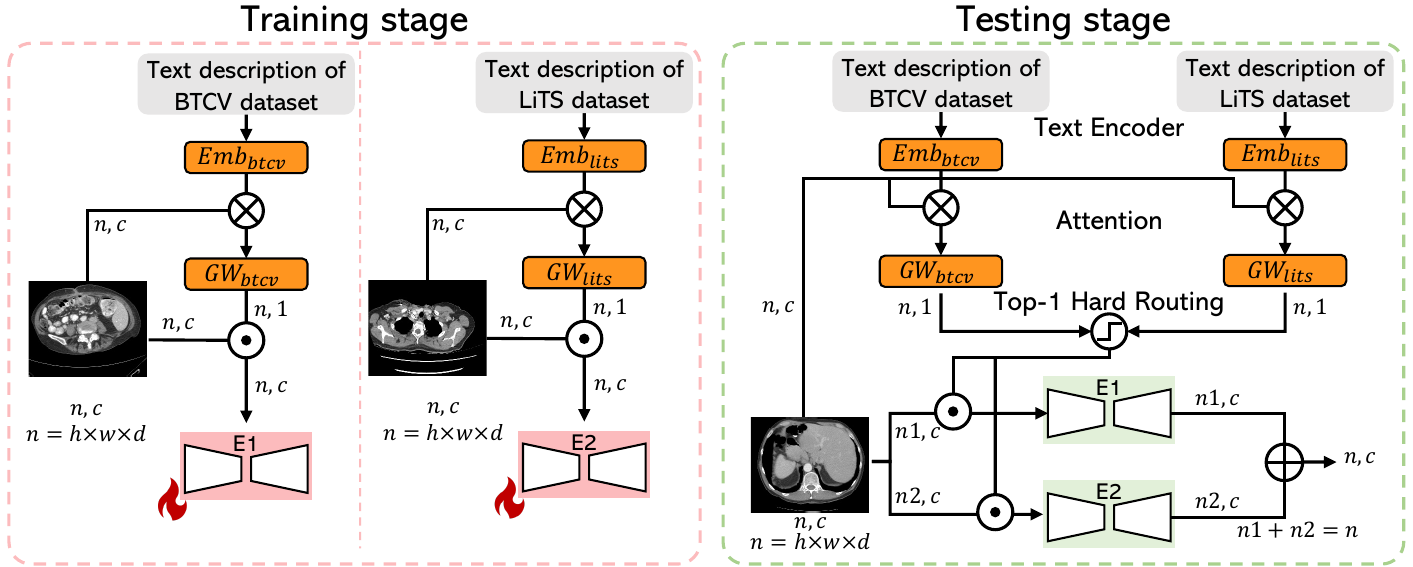}
\vspace{-2em}
\caption{Illustration of the proposed class-level gating pipeline. \textit{Emb} means Embeddings and \textit{GW} means Gating Weights. $E_1$ and $E_2$ indicate expert 1 and expert 2.}
\vspace{-1em}
\label{fig3}
\end{figure}
\vspace{-0.6cm}Specifically, we first expand the standard image-based Swin Transformer \cite{liu2021swin} into 3D CT scans. We then incorporated our MoE structure onto this enhanced backbone. 
In the training phase, as illustrated in~\ref{fig3}, we first describe the dataset $T_1$ trained in the first step (e.g.,``BTCV Segmentation dataset contains medical imaging data for abdominal organs with the following label definitions: 0.background; 1.spleen...13.left adrenal gland."). We use the text encoder from the CLIP model \cite{radford2021learning} to generate text embedding related to $T_1$. The text embedding is then matrix-multiplied with the input passed through a linear layer and a sigmoid layer to obtain a layer of language-guided gating weights (GW). GW is then multiplied by the input to obtain the parameters of $E_1$. This is the gating function of one layer of an expert. In the second step of continual learning, the parameters of the first FFN $E_1$ are fixed and loaded, and the second expert FFN $E_2$ is trained in the same gating manner.

In the testing phase, we calculate the weights for the input concerning the text embeddings of T1 and T2, respectively. For each token in the input, we only select the expert with the higher weight to perform the calculation, which is a form of top-1 hard routing. This allows each token to choose the more appropriate expert for calculation without introducing additional computational cost, thereby enabling the model to handle different categories in $T_1$ and $T_1$ simultaneously.

\section{Experimental Setup \& Result}
    \subsection{Dataset}    
    \noindent\textbf{Task-level Continual Learning}: We train and then do continual learning on the following three datasets with different orders. ACDC \cite{bernard2018deep} is a publicly-available dataset for the automated cardiac diagnosis challenge. Following \cite{lin2023convformer}, the dataset is split into 70 training samples, 10 validation samples and 20 testing samples. ISIC \cite{codella2019skin} is a publicly-availavle dataset for skin lesion segmentation, which contains 2594 images with lesions and corresponding skin lesion labels marked by experts. Among them, 162 pictures are used for testing, 162 pictures are used for verification, and 2270 pictures are used for training. COVID-19 CT Segmentation dataset \cite{fan2020inf} consists of 100 axial CT images. 
    
    \noindent\textbf{Class-level Continual Learning}: We first train on the BTCV \cite{landman2015miccai} and then do CL on the LiTS \cite{bilic2023liver}. BTCV contains 30 labeled abdominal CT scans, which we divided according to MONAI\footnote{MONAI: \url{https://monai.io/}}, with 24 for training and 6 for testing. LiTS dataset consists of 131 contrast-enhanced abdominal CT images for liver and liver tumor segmentation, originating from 7 different medical centers, which we divided into train/val/test sets according to \cite{zhang2023continual}. 

       
    \subsection{Implementation Details}

    \subsubsection{Task-level Continual Learning}: 
    We adapt the SETR \cite{zheng2021rethinking} segmentation framework for evaluating all methods, ensuring a fair comparison. 
    We employ an AdamW optimizer~\cite{loshchilov2017decoupled} for 100 epochs using a cosine learning rate scheduler with 10-epoch linear warm-up. A batch size of 16, an initial learning rate of 0.001 and a weight decay of 1e-6 are used. We use a hidden dimension of 8 for all low-rank layers. For the offline task classifier, we randomly select 8 images from each dataset as its support set and utilize the image encoder from CLIP \cite{radford2021learning} to extract their features for the task matching process. We use the most commonly used Dice score (DSC) as our evaluation metric.  
    
    \subsubsection{Class-level Continual Learning}: We use an improved Swin-UNETR \cite{hatamizadeh2021swin} implemented in MONAI as the segmentation framework. 
    We employ an AdamW optimizer~\cite{loshchilov2017decoupled} with 500 epochs for BTCV and 200 epochs for LiTS. A cosine learning rate scheduler with 10-epoch linear warm-up, a batch size of 3, an initial learning rate of 0.001 and a weight decay of 1e-5 are used. We use a hidden dimension of 8 for all low-rank layers. We use the most commonly used Dice score (DSC) as our evaluation metric. 
    Following \cite{zhang2023continual}, we report the average DSC across 13 classes from the BTCV dataset
    in step-1 learning phase and the DSC for the newly introduced liver tumor class from the LiTS dataset in the step-2 learning phase. Comprehensive implementation details are available in the Appendix.
    \subsection{Results}
    \subsubsection{Task-level Continual Learning Results.}
We first analyze the quantitative results of models on the task-level CL MIS task, as shown in Tables \ref{Tab2:task level tab2}. For more comprehensive experimental outcomes, please refer to the Appendix. The Single-task model in Tables \ref{Tab2:task level tab2} indicates results achieved by training and testing solely on the ACDC, ISIC, and COVID-19 CT datasets. The other eight rows in Tables 1 represent eight distinct scenarios, for example, ACDC $\rightarrow$ ISIC means training initially on ACDC, followed by ISIC.
The following observations can be made: firstly, the Low-Rank MoE model’s performance on both previous and current tasks is on par with or even exceeds the baseline (Single-task model), signifying that the MoE structure has effectively addressed the issue of catastrophic forgetting in CL. Secondly, compared to the baseline and models w/o low-rank architectures, those using the Low-Rank MoE structure are markedly more lightweight. Thirdly, when compared to the baseline, there is a notable improvement in the results on the current task. This improvement is attributed to the utilization of LoRA weights from previous tasks as initial values for the current LoRA weight, allowing the transfer of beneficial information from previous tasks to the current one.
\subsubsection{Class-level Continual Learning Results.}
We now analyze the quantitative results of models on the class-level CL multi-organ segmentation task, as shown in Table\ref{Tab:class-level gating}. Three single-task models refer to the results trained only on BTCV or LiTS, indicating that the results based on Swin-Tiny serve as an upper bound for all corresponding results. Compared to the other four popular CL methods, our results consistently achieve optimal performance in step 2. Qualitative result can be seen in the Appendix.
\vspace{-0.5cm}
\begin{table*}[ht]
 \renewcommand{\arraystretch}{1.0} 
  \centering
  \scriptsize
  \caption{Benchmark task-level continual learning methods. $\nabla$, $\triangle$ and $\Box$ represents ACDC, ISIC and COVID-19 CT dataset respectively. Red indicates the performance of the data trained in the final step. \#param indicates the number of trainable parameters.}
  \vspace{-1em}
  \begin{tabular}{l|c|c|c|ccc}
  \Xhline{1pt}
Model  &\#param &low-rank &MoE  &$\nabla$  &$\triangle$  &$\Box$ \\
\Xhline{0.5pt}
SOTAs &-&-& -&91.46 \cite{wu2022fat}&89.03 \cite{wu2022fat}&68.20~\cite{fan2020inf}\\
\Xhline{0.5pt}
Single-task model & 88.1M & $\times$& $\times$ & 92.02 & 90.63 &72.71\\
$\nabla \rightarrow \triangle$ & 88.1M& $\times$ & $\times$ & 50.19 & \textcolor[rgb]{1,0,0}{90.69} & -\\
$\nabla \rightarrow \Box$ & 88.1M & $\times$ & $\times$ & 32.41 &- & \textcolor[rgb]{1,0,0}{72.44}\\
$\nabla \rightarrow \triangle\rightarrow \Box$ & 88.1M & $\times$ & $\times$ &3.03 &63.60 &\textcolor[rgb]{1,0,0}{72.30}\\
$\nabla \rightarrow \Box\rightarrow \triangle$ & 88.1M & $\times$ & $\times$ &39.08 & \textcolor[rgb]{1,0,0}{90.46}&50.20\\
\hline
Single-task model & 3.4M & $\checkmark$& $\times$ & 92.08 & 90.61 &73.27\\
$\nabla \rightarrow \triangle$ & 3.4M& $\checkmark$ & $\checkmark$ & 92.08 & \textcolor[rgb]{1,0,0}{90.77} & -\\
$\nabla \rightarrow \Box$ & 3.4M & $\checkmark$ & $\checkmark$ & 92.08 &- & \textcolor[rgb]{1,0,0}{73.68}\\
$\nabla \rightarrow \triangle\rightarrow \Box$ & 3.4M & $\checkmark$ & $\checkmark$ &92.08 &90.77 &\textcolor[rgb]{1,0,0}{74.17}\\
$\nabla \rightarrow \Box\rightarrow \triangle$ & 3.4M & $\checkmark$ & $\checkmark$ &92.08 & \textcolor[rgb]{1,0,0}{90.75}&73.68\\
\Xhline{1pt}
\end{tabular}
\label{Tab2:task level tab2}
\vspace{-1.0em}
\end{table*}

\begin{table*}[ht]
\vspace{-1.2cm}
 \renewcommand{\arraystretch}{1.1} 
    \centering
    \scriptsize
    \caption{Benchmark class-level continual learning methods. $^\dag$ indicates using our improved Swin-UNETR framework. Performance for both the validation and testing sets of the LITS dataset are reported (val/test). }
    \vspace{-0.3cm}
  \renewcommand{\arraystretch}{1.0}
  \renewcommand{\tabcolsep}{2.5mm}
    \begin{tabular}{l|c|cc|cc}
        \Xhline{1pt}
        {\multirow{3}{*}{Method}} &{\multirow{3}{*}{\#param}}& \multicolumn{2}{c|}{Step1} &\multicolumn{2}{c}{Step2} \\
        \cline{3-6}
        &&BTCV &LiTS&BTCV&LiTS\\
        \cline{1-6}
        Single-task model& 62.1M & 81.9&-&-&-\\
        Single-task model$^\dag$&30.7M&82.7&-&-&-\\
        Single-task model$^\dag$&30.7M&-&56.9/49.5&-&-\\
        \hline
        LwF~\cite{li2017learning}$^\dag$&37.5M&82.7&-&76.2&49.9/43.1\\
        ILT~\cite{michieli2019incremental}$^\dag$&37.5M&82.7&-&77.8&39.0/32.8\\
        PLOP~\cite{douillard2021plop}$^\dag$&37.5M&82.7&-&78.0&41.2/36.6\\
        CLAMTS~\cite{zhang2023continual}$^\dag$&37.5M&81.9&-&78.5&50.9/45.4\\
        \hline
        Low-Rank MoE$^\dag$&2.3M&82.6&-&80.6&53.5/46.7\\
        \Xhline{1pt}
\end{tabular}
\vspace{-1.0cm}
    \label{Tab:class-level gating}
\end{table*}  

\section{Conclusion}
In this article, we propose a Low-Rank Mixture of Experts (MoE) network to address continual learning (CL) in medical image segmentation. Whenever new data arrives, the MoE structure fixes most of the parameters, only updating the data-specific expert FFN. As a result, the old data parameters are frozen within the data-specific expert, while the parameters in the new expert are activated by new data, and the parameters of the two parts of the network do not affect each other, thus resolving the catastrophic forgetting problem. To address the increased computational cost and parameter overhead in the MoE structure, we propose a low-rank decoupling parameter strategy. 
The experimental results on five public datasets demonstrate the high performance of the proposed method.
\textbf{Acknowledgments}
This work was supported in part by the Natural Science Foundation of China under Grant 82371112, 623B2001, 62394311, in part by Beijing Natural Science Foundation under Grant Z210008, in part by Shenzhen Science and Technology Program, China under Grants KQTD20180412181221912, and in part by High-grade, Precision and Advanced University Discipline Construction Project of Beijing (BMU2024GJJXK004).
\bibliographystyle{splncs04}
\bibliography{main}

\end{document}